\documentclass[runningheads]{llncs}

\usepackage{eccv}

\usepackage{eccvabbrv}
\usepackage{graphicx}
\usepackage{booktabs}
\usepackage{multirow}
\usepackage{amsmath}
\usepackage{amssymb}
\usepackage{array}
\usepackage{xcolor}
\usepackage{tikz}
\usetikzlibrary{arrows.meta,positioning,fit,backgrounds}
\usepackage{iftex}
\ifPDFTeX\usepackage[accsupp]{axessibility}\fi

\usepackage{orcidlink}
\usepackage{hyperref}

\newcommand{\maf}{Motion\-AG\-Former}

\begin{document}

\title{More Motion Is Not Always Better Motion: Corpus Composition Governs
Whether Augmentation Helps SMPL-Based Parkinsonian Gait Severity
Estimation}
\titlerunning{More Motion Is Not Always Better Motion}

\author{Michael Caiola\orcidlink{0000-0001-6355-7963} \and Andrew C. Weitz \orcidlink{0000-0002-9849-9609}}
\authorrunning{M.~Caiola and A.~Weitz}
\institute{Credence, McLean VA, USA\\
\email{mcaiola@credence-llc.com}}

\maketitle

\begin{abstract}
We grade parkinsonian gait severity (MDS-UPDRS item 3.10) from SMPL motion with an ensemble of three frozen \maf{} encoders as featurizers, reaching macro-F1 0.58 on a hidden, multi-site test set. The members differ only in the motion corpus used to pretrain them as 2D-to-3D pose lifters, so evaluating encoders singly on that test set isolates what the corpus contributes. Six pools drawn from one inertial dataset, varying only in which walking tasks they include, score between 0.32 and 0.53, and just one beats the 0.51 of an encoder given no outside motion at all. What separates them is not how much data they hold but whether they carry a contrast in walking speed, the variation this representation appears to depend on. A seventh pool that adds a third collection site (more subjects and more walks at the same task mix) scores lower still. Synthetic motion with exact ground truth fails to help for the same reason. Healthy web video fails differently: the model places it a full grade above the walks our raters scored 0, and two recovery methods agree on that placement to within 0.05. Every variant that modified the learned representation itself, rather than the corpus behind it, scored below the unmodified system.

\end{abstract}

\section{Introduction}
\label{sec:intro}

We study the estimation of the MDS-UPDRS\footnote{The International Parkinson and Movement Disorder Society (MDS)–sponsored revision of the original Unified Parkinson’s Disease Rating Scale} item 3.10 (Gait) severity
grade, an ordinal label in $\{0,1,2,3\}$, from a canonicalized SMPL
\cite{smpl} motion sequence of a single walk. The input is body motion
only, with no video, no appearance and no clinical metadata. This is the
task posed by the MoCha 2026
challenge\footnote{\url{https://mocha.care-pd.ca/challenge.html}}:
training data comes from CARE-PD \cite{carepd}, and systems are scored
on a hidden test set that pools sites absent from training.

CARE-PD spans 362 subjects -- spanning Parkinson's disease, other neurological conditions, and controls -- and 8477 walks across nine cohorts, but
only four of them, totaling 2950 walks, carry an MDS-UPDRS 3.10 gait
grade; the rest are unlabeled. That labeled pool is small and
imbalanced, so the obvious move is to bring in outside motion data. Outside
motion always arrives through some reconstruction process, and optical
capture, inertial sensing with inverse kinematics, physics-based
synthesis and monocular video recovery all emit SMPL parameters that are
routinely treated as interchangeable once they do.

We found that these sources are not interchangeable here, and, more
surprisingly, that the same source can either help or hurt depending on
which of its walks the pool contains. One inertial corpus,
converted by one pipeline, improved the model when the pool built from
it contained hurried as well as self-paced walking, and hurt the model
when it did not. What separates the two cases is whether the added
motion carries the variation the severity model depends on, which for
this representation appears to be a contrast in walking speed. Where the motion came
from and how it was recorded act through the same rule.

We report the augmentation experiments that support this, the ensemble
built on the encoder they produced, and what that ensemble is worth
against hand-crafted descriptions of the same motion; we then report
six variants that modified the learned representation rather than the
corpus behind it, all of which lost, and three external sources that
did not transfer. Most experiments are scored both offline
and on the hidden test set, and the two do not track each other because the
unseen test cohorts were captured and reconstructed differently
from those available for training.

\section{Materials and Methods}
\label{sec:methods}

\subsection{Datasets}
\label{sec:datasets}

CARE-PD \cite{carepd} aggregates nine cohorts recorded at clinical and
laboratory sites. Four carry MDS-UPDRS 3.10 severity labels over 2950
walks: BMCLab \cite{shida2023public}, PD-GaM \cite{adeli2025gaitgen},
3DGait \cite{wang2023video} and T-SDU-PD, the last introduced with
CARE-PD itself. Five are unlabeled: DNE
\cite{hoang2022towards,hoang2024smartphone}, E-LC
\cite{mckay2019freezing,kwon2023explainable}, KUL-DT-T
\cite{spildooren2010freezing,filtjens2022automated}, and T-SDU and
T-LTC, also introduced with CARE-PD.

Every lifting encoder starts from Human3.6M \cite{h36m}, the
standard pose-lifting pool, and augments it with one of two outside
corpora. BEDLAM \cite{bedlam} is a large
synthetic dataset whose motion is authored rather than estimated, so its
SMPL parameters are exact by construction. WearGait-PD \cite{weargait}
is a corpus of walking recorded with body-worn inertial
sensors.

WearGait-PD was collected under one protocol and one set of equipment
at three sites: two Johns Hopkins campuses (JHOC and Bayview) and the VA
Puget Sound Center for Limb Loss and Mobility (VA-Seattle). Except where
noted, every WearGait result here draws on the two Johns Hopkins sites
only, giving 120 subjects (50 control, 70 Parkinson's) who contribute
349 walks of self-paced walking, hurried-paced walking and
timed-up-and-go. VA-Seattle was added in a later attempt, bringing the
total WearGait pool to 178 subjects (80 control, 98 Parkinson's) and 523
walks at the same task composition.

Two further sources were tested as severity training data rather than as
lifting corpora. From a public YouTube-sourced gait dataset \cite{gavd}
we reconstructed 271 clips labeled normal and 43 labeled parkinsonian
with two monocular recovery methods, WHAM \cite{wham} and GVHMR
\cite{gvhmr}. To compare the two where the ground truth is certain, we
additionally recorded five volunteers on a smartphone, each walking once
unaided and once pushing an office chair, and reconstructed all ten of
those walks both ways as well. We also rendered
SMPL meshes for 183 walks from the five unlabeled CARE-PD cohorts,
graded them manually against the MDS-UPDRS 3.10 rubric \cite{updrs}, and
added them to training. All grading was done by a single non-clinician
rater.

\subsection{Inertial-to-SMPL conversion and lifting corpora}
\label{sec:corpora}

WearGait recordings are converted to SMPL by fitting the per-frame
sensor orientations to the SMPL kinematic tree with inverse kinematics:
each sensor's orientation is aligned to a still-window calibration from
the start of the trial, mapped onto its SMPL bone, and converted to the
body model's local joint rotations. Translation is not recoverable from
inertial data alone and is set to zero, so each sequence carries pose
only. The 349 walks of the two-site pool expand to 698 training
sequences through two temporal downsampling rates and two camera
azimuths.

Four lifting encoders differ only in their training corpus.
\texttt{liftbase} uses the standard pose-lifting pool alone,
\texttt{liftbedlam} adds BEDLAM, \texttt{liftwg} adds the WearGait pool
described above, and \texttt{liftwglong} trains on the \texttt{liftwg} pool
with a longer schedule.

The pool named \texttt{liftwg} above is one of six we built from the
WearGait recordings, four of them keeping whole trials and two cutting
the same trials into short clips, and the six differ in which walking
tasks they contain. Of the whole-trial pools, \texttt{liftwg} (120
subjects) is the one used in the final system; \texttt{liftwgmore}
adds further walks from those same 120 subjects, \texttt{liftwgall} adds
35 more subjects (155 in all), and \texttt{liftwgsp} keeps only the
self-paced walks. The two clip pools are cut at matched task composition
from the recordings behind \texttt{liftwgsp} and \texttt{liftwg}, giving
\texttt{liftspseg} and \texttt{liftmixseg} respectively. Counts for each
pool are given with the results in Table~\ref{tab:pools}.

A seventh pool adds the third collection site, keeping the same three
tasks in the same proportion as the two-site pool (176 self-paced, 176
hurried-paced and 171 timed-up-and-go, against 118, 118 and 113 before)
and holding the conversion pipeline, the encoder architecture and the
training schedule fixed.

\subsection{Severity model}
\label{sec:model}

Each SMPL sequence is rendered to 2D joint projections from two
synthetic camera views and cut into overlapping 81-frame clips, matching
the input convention of \maf{} \cite{motionagformer}. A frozen \maf{}
encoder maps each clip and view to a 512-dimensional embedding, and the
embeddings are averaged over clips and views to give one descriptor per
walk. The encoder never sees severity labels, since it is trained as a
2D-to-3D lifting model and only the head is supervised. The lifting
corpus is therefore the only thing that varies across the encoders
compared in Section~\ref{sec:liftresults}.

\begin{figure}[h!]
\centering
\resizebox{\textwidth}{!}{%
\begin{tikzpicture}[
  font=\scriptsize,
  box/.style={draw, rounded corners=2pt, minimum height=6mm, align=center,
              inner sep=2pt},
  enc/.style={box, fill=blue!6, minimum width=18mm},
  hd/.style={box, fill=gray!8, minimum width=14mm},
  ar/.style={-{Latex[length=1.6mm]}, semithick}
]
\node[box, minimum width=14mm] (smpl) {SMPL\\sequence};
\node[box, right=4mm of smpl, minimum width=16mm] (proj) {2-view\\projection};
\node[box, right=4mm of proj, minimum width=16mm] (clip) {81-frame\\clips};

\node[enc, right=7mm of clip, yshift=9mm]  (e1) {small\\26$\times$64};
\node[enc, right=7mm of clip]              (e2) {large\\26$\times$128};
\node[enc, right=7mm of clip, yshift=-9mm] (e3) {large\\26$\times$128, seed 2};

\node[hd, right=5mm of e1] (h1) {MLP head};
\node[hd, right=5mm of e2] (h2) {MLP head};
\node[hd, right=5mm of e3] (h3) {MLP head};

\node[box, right=6mm of h2, fill=green!8, minimum width=14mm] (avg)
  {mean\\softmax};
\node[box, right=4mm of avg, minimum width=10mm] (out) {grade\\$0$--$3$};

\draw[ar] (smpl) -- (proj);
\draw[ar] (proj) -- (clip);
\foreach \e in {e1,e2,e3} \draw[ar] (clip.east) -- (\e.west);
\draw[ar] (e1) -- (h1);
\draw[ar] (e2) -- (h2);
\draw[ar] (e3) -- (h3);
\foreach \h in {h1,h2,h3} \draw[ar] (\h.east) -- (avg.west);
\draw[ar] (avg) -- (out);

\begin{scope}[on background layer]
\node[draw, dashed, rounded corners=3pt, inner sep=3.5mm,
      fit=(e1)(e3), label={[font=\tiny]above:frozen, lifting-trained}] {};
\end{scope}
\end{tikzpicture}%
}
\caption{Inference path. A walk's SMPL sequence is rendered to 2D joint projections from two synthetic camera views, cut into 81-frame clips, and embedded by three frozen \maf{} encoders. Embeddings are averaged over clips and views, standardized, and passed to per-member MLP heads whose softmax outputs are averaged. The encoders share one WearGait-augmented lifting corpus and differ only in backbone size and seed.}
\label{fig:system}
\end{figure}
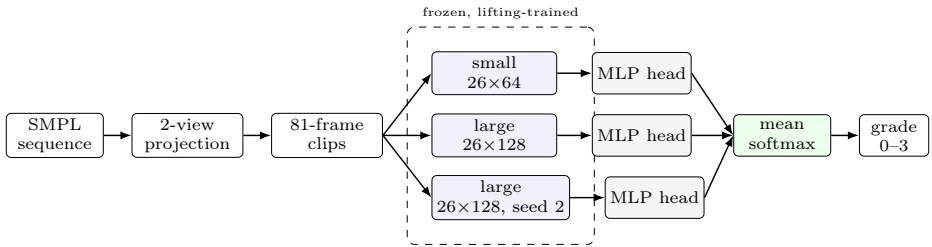

Each ensemble member stores the mean and the standard deviation of
its own training features and uses them to standardize test features. The
standardized descriptor is passed to an MLP of widths
$512 \rightarrow 256 \rightarrow 128 \rightarrow 4$, with batch
normalization after each hidden layer and dropout 0.5, trained with
class-weighted cross-entropy. At inference the per-member softmax
outputs are averaged, at the weights given in
Section~\ref{sec:enscomp}, and the arg-max is taken. Members differ only in
\maf{} backbone size and random seed (Fig.~\ref{fig:system}).

As an alternative to learned features, we evaluated two hand-crafted
pipelines over the same SMPL motion: 434 statistics of the joint
rotations (range, variability and spectral content of each joint angle,
harmonized across sites before a classifier is fitted), and 26 clinical
spatiotemporal gait parameters (cadence, step length, stride-time
variability, walking speed), late-fused with the frozen-encoder
ensemble.

\subsection{Ensemble compositions and test-time centering}
\label{sec:enscomp}

With the lifting corpus fixed, the remaining design choice is the
backbone. We built compositions from four \maf{} sizes, written here as
layers$\times$width: small (26$\times$64), base (16$\times$128), large
(26$\times$128) and xlarge (40$\times$128). Each size was evaluated on its
own and in ensembles of two, three and four members.

One composition, a three-member ensemble of one small and two large
encoders, is the final system released with this paper, with the members
averaged at $0.4/0.3/0.3$ to upweight the WearGait-augmented small
member. Section~\ref{sec:ensemble} compares it against every other
composition we evaluated.

\label{sec:siteclus}%
The test set pools sites absent from training, and site is a large
source of feature variance, so we tested transductive alignment, which
recenters test features on test-set statistics. Because pooling over a
mixed-site test set recenters on a mean that belongs to no site, we also
built a per-site variant that first groups test walks by a fingerprint
correlated with site but not with severity (the capture-rate group
crossed with a straight-versus-turning split of the pelvis trajectory)
and recenters within each group, falling back to training statistics for
walks that are not confidently attributed.

\subsection{Evaluation protocol}
\label{sec:eval}

Offline results are leave-one-cohort-out (LOCO) over the four labeled CARE-PD
cohorts (2950 walks) at three seeds. The offline metric is macro-F1 over
classes 0, 1 and 2, because class 3 appears in only two of the four
labeled cohorts, so a LOCO split either trains or tests
without it. Hidden-test results are macro-F1 over all four classes on
the held-out multi-site test set of the MoCha 2026 challenge. The test
set, its labels and any per-walk feedback were never available to us:
every hidden-test number reported here comes from a single evaluation of
that configuration through the challenge server.

Two further measurements support the interpretation. To ask how much of
the severity signal is carried by walking speed, we repeated 0-versus-1
discrimination on the labeled cohorts with walks matched for speed, and
ran the same test on three other representations. The second
measurement required additional gait grading: videos of body
meshes were rendered from the SMPL parameters, watched, and assigned a
grade of 0 to 3 manually against the MDS-UPDRS 3.10
rubric \cite{updrs}, with no real video and no clinical information
available. All 183 walks of the five unlabeled CARE-PD
cohorts were graded this way and added to training as if they carried
clinical labels. To ask what such a grade is worth, we ran two blind
calibration rounds in which walks from the \emph{labeled} cohorts were
graded by the same rater under the same conditions with the clinical
labels withheld, then scored against those labels. The first round
covered 46 walks spanning all four classes; the second, run after the
first had located the rater's difficulty, covered 45 walks balanced over
classes 0--2.

\section{Results}
\label{sec:results}

\subsection{Lifting-corpus composition}
\label{sec:liftresults}

Table~\ref{tab:ablation} reports the four lifting encoders under both
protocols. WearGait augmentation improves every offline column, and the
longer schedule improves it further. BEDLAM does the opposite, falling
below the unaugmented baseline on macro-F1 and hardest of all on class-0
recall. Seed spread is small relative to the gaps, so the offline
ranking is stable, but these are four points in a large design space,
and we did not sweep augmentation ratio and schedule length jointly.

\begin{table}[h!]
\centering
\caption{Lifting-corpus ablation. Offline columns: LOCO over the four
labeled CARE-PD cohorts (2950 walks), three seeds, one ensemble member;
macro-F1 over classes 0/1/2 because class 3 appears in only two labeled
cohorts. Last column: four-class macro-F1 on the hidden test, one
evaluation of the encoder alone. Only the lifting corpus differs
between rows.}
\label{tab:ablation}
\small
\setlength{\tabcolsep}{4pt}
\begin{tabular}{lccccc}
\toprule
 & macro-F1 & class-0 & class-0 & & hidden \\
Lifting encoder & (0/1/2) & F1 & recall & acc. & test \\
\midrule
liftbase \footnotesize(no augmentation)    & 0.374 $\pm$ 0.005 & 0.495 & 0.577 & 0.410 & 0.51 \\
liftbedlam \footnotesize(+ synthetic)      & 0.363 $\pm$ 0.006 & 0.401 & 0.459 & 0.378 & 0.43 \\
liftwg \footnotesize(+ WearGait)           & 0.393 $\pm$ 0.001 & 0.510 & 0.640 & \textbf{0.422} & \textbf{0.53} \\
liftwglong \footnotesize(+ WearGait, long) & \textbf{0.406} $\pm$ 0.006 & \textbf{0.528} & \textbf{0.658} & 0.419 & 0.47 \\
\bottomrule
\end{tabular}
\end{table}

The hidden test, which set the challenge ranking and is the only
multi-site held-out measurement available, gives a partly different
picture. WearGait at
the standard schedule still improves on the unaugmented encoder, and
BEDLAM is still worst, by a wide margin. But the longer WearGait
schedule, the best of the four offline, is second worst on the test
set, an instance of the validation problem of
Section~\ref{sec:limits}.

\begin{figure}[h!]
\centering
\includegraphics{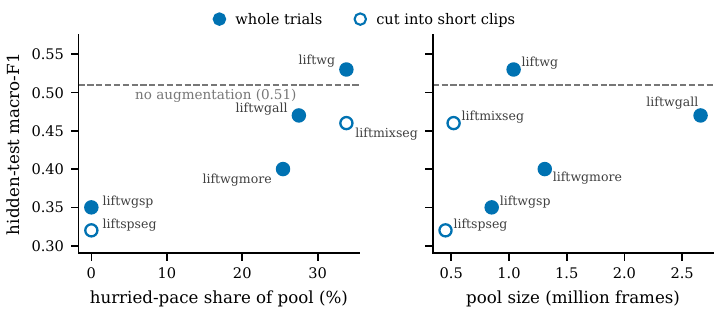}
\caption{The six WearGait pools of Table~\ref{tab:pools} against the
hidden test. Left: hidden-test macro-F1 against the pool's share of
hurried-pace walking; among the four whole-trial pools, more
hurried-pace walking means a higher score. Right: the same six pools
against corpus size, which shows no such pattern; the largest pool
stays below the no-augmentation baseline. One evaluation per pool, each
as a single encoder.}
\label{fig:pools}
\end{figure}

The six WearGait pools isolate task composition: all share the same
recordings, sensors and conversion, and differ only in which walking
tasks they contain (Table~\ref{tab:pools}, Fig.~\ref{fig:pools}). Only
one beats the unaugmented encoder and the other five score below it, so
the same recordings both help and hurt depending on their task mix.
Among the four whole-trial pools, the more hurried-pace walking a
pool contains, the higher it scores, a perfect rank match, while corpus
size gives no such pattern: the largest pool, at 2.5 times the frames
and 35 more subjects than the selected one, scores 0.06 lower. Cutting
the same recordings into short clips costs roughly 0.07 at matched task
composition, a second and smaller factor.

\begin{table}[h!]
\centering
\caption{WearGait pool composition against the hidden test. The six
top rows share the same recordings, sensors and inertial-to-SMPL
conversion, differing in walking tasks and in whole trials versus short
clips. Scores are hidden-test macro-F1, one evaluation per pool, each
as a single encoder (not comparable with ensemble figures elsewhere),
against 0.51 unaugmented. \texttt{liftwg} is the pool in the final
system. The last row adds a third collection site at fixed task
composition; $\dagger$ its score is the three-member ensemble, as the
single encoder was not evaluated.}
\label{tab:pools}
\small
\begin{tabular}{lrrrrcc}
\toprule
Pool & Subjects & Sequences & Frames & Hurried-pace & Whole trials
& Hidden-test \\
     &          &           &        & share        & or clips
& macro-F1 \\
\midrule
liftspseg  & 119 & 3012 & 0.45\,M & 0\%    & clips & 0.32 \\
liftwgsp   & 119 &  692 & 0.85\,M & 0\%    & whole & 0.35 \\
liftwgmore & 120 &  928 & 1.31\,M & 25.4\% & whole & 0.40 \\
liftmixseg & 120 & 3246 & 0.52\,M & 33.8\% & clips & 0.46 \\
liftwgall  & 155 & 1716 & 2.66\,M & 27.5\% & whole & 0.47 \\
liftwg     & 120 &  698 & 1.04\,M & 33.8\% & whole & \textbf{0.53} \\
\midrule
liftwg + 3rd site & 178 & 1046 & --- & 33.8\% & whole & $0.50^\dagger$ \\
\bottomrule
\end{tabular}
\end{table}

The last row of Table~\ref{tab:pools} separates volume from composition
directly: it adds the third collection site to the selected pool,
raising it from 120 to 178 subjects and 349 to 523 walks at fixed task
proportions and pipeline (Section~\ref{sec:corpora}). LOCO macro-F1 over
classes 0/1/2 falls from 0.370 to 0.356 and class-0 recall from 0.625 to
0.474, and the three-member ensemble built on it scored 0.50 against
0.58. The extra subjects and the extra site therefore cost more than any
change of task composition, with the loss again falling on class-0
recall, as with BEDLAM. This is the only comparison reported here that
varies corpus size without also varying what the corpus contains.

Severity prediction from this representation tracks walking speed
closely. On the labeled cohorts, 0-versus-1 discrimination falls from
0.63 to 0.47 AUC, at or below chance, once walks are matched for speed,
and the three other representations tested the same way land at chance
as well.

Placing each encoder inside the three-member ensemble weakens the
picture (Table~\ref{tab:ensmember}): seed spread widens threefold, the
augmented and unaugmented variants no longer separate (0.365 against
0.391, about one and a half standard deviations, in the reversed
order), and only class-0 recall keeps its standalone ranking. Every
lifting member still beats having none. We therefore state the
augmentation result at the level of the single encoder and of class-0
recall, and claim no ensemble-level macro-F1 gain; the ensemble
measurement is inconclusive at this spread rather than negative.

\begin{table}[h!]
\centering
\caption{Lifting encoder inside the three-member ensemble.
LOCO over the four labeled CARE-PD cohorts, three seeds,
macro-F1 over classes 0/1/2. The two large members and the rest of the
pipeline are held fixed, and only the small member varies. Offline
only: not every small member was evaluated on the hidden test inside
this ensemble.}
\label{tab:ensmember}
\begin{tabular}{lccc}
\toprule
Small member & macro-F1 (0/1/2) & class-0 F1 & class-0 recall \\
\midrule
liftbase          & \textbf{0.391} $\pm$ 0.011 & \textbf{0.496} & 0.599 \\
liftbedlam        & 0.362 $\pm$ 0.019 & 0.430 & 0.490 \\
liftwg            & 0.365 $\pm$ 0.014 & 0.465 & 0.609 \\
liftwglong        & 0.372 $\pm$ 0.012 & 0.486 & \textbf{0.621} \\
no lifting member & 0.357 $\pm$ 0.020 & 0.411 & 0.519 \\
\bottomrule
\end{tabular}
\end{table}

\subsection{Ensemble composition}
\label{sec:ensemble}

Where the corpus enters the pipeline matters as much as what it
contains. Supplying the same WearGait recordings to the heads as
labeled severity training data, rather than to the encoder as lifting
material, scored 0.54 against the 0.58 that the final three-member
ensemble reaches on the hidden test. The
recordings that improve a lifting encoder therefore degrade a severity
head, and the gain reported above is a property of how the corpus is used
and not of the corpus alone.

Table~\ref{tab:ensemble} gives hidden-test macro-F1 for every
composition we evaluated, grouped by whether the small encoder is
present.

\begin{table}[h!]
\centering
\caption{Ensemble composition against the hidden test. Learned-feature
members are lifting-trained on the same WearGait-augmented corpus and
differ in backbone size and random seed; one evaluation per composition.
Sizes are layers~$\times$~width; xlarge is 40 layers. The last
block replaces the learned features with hand-crafted descriptors.}
\label{tab:ensemble}
\small
\begin{tabular}{llc}
\toprule
& composition & hidden-test macro-F1 \\
\midrule
\multirow{2}{*}{single}
  & base ($16\times128$)                    & 0.51 \\
  & small ($26\times64$, \emph{liftwg})     & 0.53 \\
\midrule
\multirow{2}{*}{pairs without the small encoder}
  & base $+$ large                          & 0.51 \\
  & large $+$ xlarge                        & 0.52 \\
\midrule
\multirow{2}{*}{pairs with the small encoder}
  & small $+$ base                          & 0.54 \\
  & small $+$ large                         & 0.56 \\
\midrule
three members
  & small $+$ large $+$ large (seed 2)      & \textbf{0.58} \\
\midrule
\multirow{2}{*}{four members}
  & $+$ second small (seed 2)          & 0.55 \\
  & $+$ separate-corpus member         & 0.56 \\
\midrule
\multirow{2}{*}{hand-crafted}
  & 434 joint-rotation statistics      & 0.41 \\
  & 26 gait parameters (late-fused)    & 0.34 \\
\bottomrule
\end{tabular}
\end{table}

Capacity alone does not explain the ranking: the wider base backbone scores
below the small \emph{liftwg} encoder, the two pairs built without that
encoder score at or below what it reaches on its own, and the 40-layer
variant lowered each composition we paired it with. What helps is
pairing the small encoder with a larger one and adding a third member of
a different seed; a fourth member costs roughly 0.02, observed twice
with different choices of fourth member. The small 26$\times$64 encoder
is therefore load-bearing rather than one interchangeable member of
three.

The final system reaches macro-F1 $0.58$ with precision $0.65$, recall
$0.55$ and accuracy $0.53$. Its $0.4/0.3/0.3$ weighting and an
equal-weight average score identically on every one of those metrics and
separate only on quadratic weighted kappa, by a single point ($0.41$
against $0.40$), which we read not as evidence that the weighting
matters but as evidence that redistributing weight across members cannot
meaningfully move macro-F1.

Transductive centering does not help either. Pooled over the whole
test set it scored 0.49 and 0.51 against the 0.58 baseline, collapsing
precision rather than recall; the per-site variant of
Section~\ref{sec:siteclus}, forced to engage on every walk, scored
0.52. Recentering features per recovered site therefore costs what the
other feature-space modifications cost.

Taken with the four-member and forty-layer results, this flatness
suggests that $0.58$ is a stable ceiling for this family of
frozen-encoder ensembles rather than a point that further ensemble
engineering will improve.

Replacing the learned features with hand-crafted ones costs far
more than any composition choice (Table~\ref{tab:ensemble}, last block):
434 SMPL joint-rotation statistics with site harmonization score 0.41,
and 26 spatiotemporal gait parameters late-fused with the encoder score
0.34, against 0.58 for the frozen-encoder ensemble they were built to
replace and to reinforce.

\subsection{Changes to the learned representation}
\label{sec:repchange}

Two variants altered the frozen representation itself rather than the
corpus behind it or the heads on top of it. The first expanded the
synthetic camera rig from two views to six, sampling the subject from
both sides rather than only the right. The second unfroze the top two of
the 26 \maf{} blocks and fine-tuned them jointly with the head at a
learning rate one hundred times lower. Offline both looked like our best
candidates, and both improved class-0 recall.

Both scored 0.52 on the hidden test, against 0.58 for the system they
modified; per-site recentering also scored 0.52, full unfreezing scored
0.40 and 0.35, and every system that left the features untouched
returned 0.58 (Table~\ref{tab:repchange}). Changing the lifting corpus
moves the representation too: the three-site encoder's features sit at
cosine 0.77 to those of the member it replaces, against 0.93 to 0.96 for
the view change. The table ranks these by how much of the
representation each one altered, and the score falls monotonically with
that amount.

\begin{table}[h!]
\centering
\caption{How much of the frozen representation each variant changed,
against its hidden-test score. Offline gains are LOCO
over the four labeled CARE-PD cohorts, three seeds, macro-F1 over
classes 0/1/2. Hidden-test scores are for the full three-member
ensemble with the change applied, so 0.58 is the unmodified final
system.}
\label{tab:repchange}
\small
\begin{tabular}{llcc}
\toprule
extent of change & variant & offline gain & hidden test \\
\midrule
none & frozen encoders            & ---      & \textbf{0.58} \\
\midrule
features recentered & per-site, forced   & ---      & 0.52 \\
 & pooled over the test set                        & ---      & 0.49--0.51 \\
input geometry & six camera views          & $+0.044$ & 0.52 \\
top 2 of 26 blocks & partial unfreeze      & $+0.030$ & 0.52 \\
lifting corpus & three-site, 178 subjects & $-0.014$ & 0.50 \\
\midrule
all 26 blocks & end-to-end fine-tuning     & ---      & 0.40 \\
 & end-to-end, stronger regularization      & ---      & 0.35 \\
\bottomrule
\end{tabular}
\end{table}

\subsection{External sources that did not transfer}
\label{sec:external}

The mesh grading of the unlabeled cohorts looked positive
offline and scored 0.52 against 0.58 on the hidden test. The blind
calibration puts a number on the labels it supplied. In the four-class
round, agreement with the withheld clinical labels is 0.46 exact and
0.93 within one class, but the four-class macro-F1 of the grading is
0.358, below the 0.58 of the model it was meant to improve, and every
class-3 walk was graded 2. Restricted to classes 0--2, the same
rater reaches 0.53 exact and macro-F1 0.537, so the four-class deficit
is concentrated in class 3, which a mesh cannot express.

\begin{figure}[h!]
\centering
\includegraphics{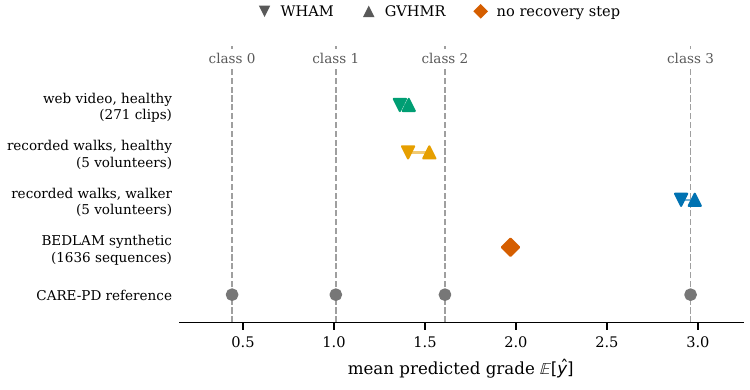}
\caption{Where external motion sources land in the final system's
predicted-grade space; the dashed lines are its mean predicted grade on
CARE-PD walks of classes 0--3 ($0.44$, $1.01$, $1.61$, $2.96$). Each
source appears under both recovery methods, joined by a line. Every
healthy external source sits about a grade above the class-0 reference,
and the spread between recovery methods within a source ($0.05$ to
$0.12$) is a fraction of that offset. The recorded walker walks land at
the class-3 reference, matching the assistive-device criterion from
kinematics alone. BEDLAM carries the organizers' own ground-truth SMPL
and has no recovery step.}
\label{fig:placement}
\end{figure}

Web video also failed to transfer, but not for reasons of
reconstruction. Reconstructing the same 271 healthy web clips both ways
moves their mean predicted grade only from 1.36 (WHAM) to 1.41 (GVHMR),
and on the ten recorded walks, where the two reconstructions describe
the same physical walk, the methods agree to within a tenth of a grade
(mean absolute difference 0.11 on a 0--3 scale, maximum 0.33). Recovery
choice between two current methods is not a material source of variance
for this system. The offset that remains is much larger. Every healthy
source we placed sits about a grade above the CARE-PD walks its raters
scored 0 (0.44): the web clips at 1.36 and 1.41, the volunteers' unaided
walks at 1.41 and 1.52, whichever method reconstructed them
(Fig.~\ref{fig:placement}). The ordering, by contrast, survives: over the
40 parkinsonian clips the rater graded, predicted grade rises with the
rater's grade at Spearman $\rho = 0.67$ under WHAM and $0.64$ under
GVHMR ($p < 10^{-5}$ for both). The walker walks land at 2.91
and 2.98, matching the class-3 criterion (assistive-device use) from
kinematics alone, with no walker present in the SMPL input.

Adding the 271 healthy clips as class-0 training examples confirms
the placement reading: LOCO macro-F1 over classes 0/1/2 falls from
0.363 to 0.311, and upweighting the clips threefold drops it further to
0.282. So the clips are unusable as training material for that class,
though not because their poses are wrong: the model simply does not
score them where their label says they belong.

\section{Discussion}
\label{sec:discussion}

\subsection{What governs whether augmentation helps}
\label{sec:governs}

The six pools of Table~\ref{tab:pools} share one corpus and one
conversion pipeline, so where the motion came from is held fixed across
them, and they still span 0.32 to 0.53 against 0.51 for adding nothing
at all. The three-site pool holds task composition fixed while adding a collection site and 58 more subjects, and it loses more than any composition variant, so volume does not act in the expected direction over the range we can reach. Read with the speed-matched AUC
result, how the pools rank points to a single rule. Augmentation helps
when the added motion contains the variation the task depends on, and
this representation depends on contrast in walking speed. An encoder can
only represent a speed-dependent quantity well if its training pool
contains speed contrast, and self-paced walking alone gives it none.
With four whole-trial pools from one dataset we cannot separate the
hurried-pace fraction from the other task content that replaces it, so we
report how the pools rank and the mechanism it suggests rather
than an isolated cause.

The source of the motion acts through this same rule. BEDLAM is
larger than WearGait and its SMPL parameters are exact by construction,
so both a volume account and a label-quality account predict that it
should help. Instead it hurts, behaving like the WearGait pools built
without a speed contrast.

External healthy motion is a third instance of the same rule
(Section~\ref{sec:external}). Its reconstructions are consistent across
recovery methods and the model orders its parkinsonian clips correctly,
but every healthy external source we placed, web clips and our own
recordings alike, sits about a grade above the walks CARE-PD's raters
scored 0. A grade of 0 on item 3.10 means normal gait, so those walks
and ordinary healthy walking should be the same thing, and the model
does not treat them as such. We cannot say from these measurements why this is the case. Walking speed is the obvious candidate, given how
closely severity tracks it here, and speed is exactly what differs
between a timed clinical walk and someone crossing a car park on
YouTube. Other possible candidates include capture geometry and age. Whatever the cause,
it sits between the labeled corpus and everything outside it, and it is
why external healthy motion cannot supply the class-0 examples the
labeled pool lacks.

Reconstruction is an issue inside CARE-PD as well: its cohorts were
captured by different modalities (multi-camera RGB, optical motion
capture, ceiling-mounted video) and fitted by different pipelines, so a
model trained on it must generalize across reconstruction methods as
well as sites, which is part of why a split over four such cohorts
predicts the test set poorly.

Part of the mesh-grading failure is definitional rather than a matter
of rater skill. Grade 3 is defined by the use of a walking aid, and an
aid is not part of the SMPL body model, so that information is absent
from the rendered mesh rather than degraded in it, and no amount of care
by the rater recovers it: in the blind calibration, every class-3 walk
was graded 2. estricted to classes 0--2 the same rater's grading
is reliable (macro-F1 0.537, Section~\ref{sec:external}), so the whole
deficit is class 3.

\subsection{Why the representation resists improvement}
\label{sec:resists}

Six variants altered the frozen representation, by recentering its
features, by changing the camera geometry that produces them, or by
training the encoder that emits them. All six scored between 0.35 and
0.52, and the loss grows with how much of the representation was
altered (Table~\ref{tab:repchange}). Two of them were our strongest
offline candidates.

Which class absorbed the loss cannot be read from aggregate scores,
but class 3 is the likeliest. It is defined by the use of a walking aid,
invisible in the mesh, and within a cohort walking speed alone separates
it with high AUC -- so the model appears to hold it at an
extreme of the feature space rather than behind a boundary drawn against
its neighbors, and a class held at an extreme is the first thing a
perturbation of that space disturbs.

No offline instrument we had could have caught the loss:
leave-one-cohort-out cannot measure class 3 at all
(Section~\ref{sec:limits}), and holding out one class-3 subject at a
time saturates near the ceiling for every configuration we tried. We
therefore report the pattern as a constraint rather than a prediction:
on this task, changes that preserve the representation can be evaluated
offline, and changes that alter it cannot.

\subsection{Limitations}
\label{sec:limits}

\begin{figure}[h!]
\centering
\includegraphics{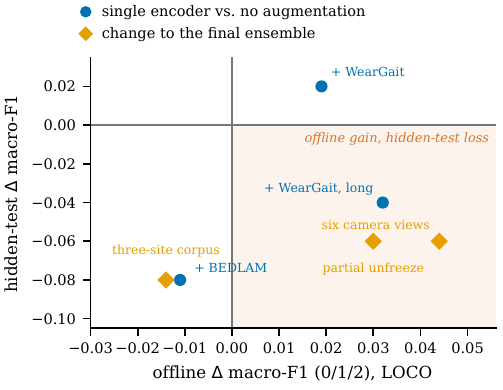}
\caption{Offline movement against hidden-test movement for the six
experiments measured under both protocols. Circles are single lifting
encoders against the unaugmented baseline (offline: LOCO macro-F1 over
classes 0/1/2; hidden test: single-encoder evaluations against 0.51).
Diamonds are changes to the final ensemble (offline gains as in
Table~\ref{tab:repchange}; hidden test against 0.58), so deltas are
comparable within a marker family rather than across the two. Offline
selection would have picked the two largest offline gains; both lost.}
\label{fig:offboard}
\end{figure}

Leave-one-cohort-out did not reliably predict hidden-test movement, so
the offline tables here are internal comparisons under a fixed protocol,
not estimates of test performance. Fig.~\ref{fig:offboard} plots the six
measurements for which we hold both deltas; the clearest instance is
the six-view rig, which improved offline macro-F1 over classes 0/1/2 by
0.044, among our largest offline gains, and lost 0.06 on the hidden
test. Four cohorts are too few domains to average over, and they do not
span the site variation of the test set. The volume comparisons are also narrow: the WearGait pools span 0.45 to 2.66 million frames and 119 to 178 subjects, so they show volume failing to help over a roughly sixfold range, not that scale can never help; an orders-of-magnitude larger pool of the right composition is untested.

Class 3 has 44 walks from 7 subjects in two of the four labeled cohorts.
A leave-one-cohort-out split therefore either trains or tests without
it, so no held-out estimate of class-3 performance is meaningful here.
That is why the offline metric is macro-F1 over classes 0, 1 and 2 while
the hidden-test metric covers all four, and the two are comparable
within columns rather than in level.

The recovery-method comparison covers two methods over two sources:
271 web clips reconstructed both ways, and ten walks from five
volunteers filmed once and reconstructed twice. It bounds the recovery
sensitivity of this system on this footage
(0.05 grades between reconstructions of the same clips, 0.11 between
reconstructions of the same walk), not of SMPL pipelines in general. It
does not identify what else moves the placement of external healthy
walks, only that the recovery step is too small to account for it.
All grading reported here was done by one non-clinician rater: the
183 mesh renderings from the five unlabeled CARE-PD cohorts, the 91
walks of the two blind-calibration rounds, and the parkinsonian clips
from the web-video
source, which carry no clinical grade of their own. The blind
calibration measures that rater against clinical labels; its second
round separates the definitional class-3 failure from the remaining
classes, but within classes 0--2 it still cannot separate rater skill
from the limits of the mesh representation. The hidden-test comparisons in Section~\ref{sec:ensemble} are single
measurements on one test set, so small differences between adjacent rows
are not resolved.

\section{Conclusion}
\label{sec:conclusion}

Our system grades MDS-UPDRS gait severity from SMPL motion with three
frozen \maf{} encoders of different sizes, lifting-trained on one
inertially-derived corpus, feeding per-member heads whose softmax
outputs are averaged at fixed weights. It reaches macro-F1
0.58 on a hidden multi-site test set,
against 0.51 for the same architecture without outside
motion and 0.53 for its strongest single member. Hand-crafted
descriptors of the same SMPL motion fall much further behind, at 0.41
and 0.34, so the learned representation is worth roughly 0.17 macro-F1
over hand-crafted descriptions of the motion it is computed from.

Two findings came out of building it. The first concerns what to add.
Across every corpus we tested, one variable predicted whether adding
motion helped, and it was not volume: pools carrying a contrast in
walking speed improved the encoder, and pools without one, however
large and however exact their ground truth, scored below adding nothing
at all. Healthy web video is misplaced by a full grade under either recovery
method while its parkinsonian clips still come out in the right order,
which makes such footage readable for relative severity and useless for
calibration; and the recordings that improve the
encoder as lifting material hurt the heads when supplied as labels. What the added
motion contains, and where it enters the pipeline, matter; how much of
it there is does not.

The second concerns what not to touch. Every variant that modified the
frozen representation lost, by an amount that grew with how much it
modified, and no offline instrument we had could see the damage in
advance. The likeliest casualty is class 3, which the model appears
to hold at an extreme of its feature space, where a perturbation strikes
first.

\subsubsection{Code and model availability.}
The final system, trained heads and encoder checkpoints, and the
corpus-building scripts are available at
\url{https://github.com/mikecredence/mocha-challenge}. CARE-PD, WearGait-PD, BEDLAM and
GAVD are available from their original authors under their own terms.

\bibliographystyle{splncs04}
\bibliography{main}

\end{document}